\documentclass{article}

\usepackage[margin=37mm]{geometry}
\pdfoutput=1
\usepackage[utf8]{inputenc} 
\usepackage[T1]{fontenc}    
\usepackage{hyperref}       
\usepackage{url}            
\usepackage{booktabs}       
\usepackage{amsmath}
\usepackage{amsfonts}       
\usepackage{nicefrac}       
\usepackage{microtype}      
\usepackage{amssymb}
\usepackage{afterpage}
\usepackage{multicol}
\usepackage{graphicx}
\usepackage[caption=false]{subfig}
\usepackage{xcolor}
\usepackage{algorithm}
\usepackage[noend]{algpseudocode}
\usepackage{adjustbox}

\newcommand{\norm}[1]{\left\lVert#1\right\rVert}

\newcommand*\xbar[1]{%
\hbox{%
\vbox{%
\hrule height 0.5pt 
\kern0.5ex
\hbox{%
\kern-0.2em
\ensuremath{#1}%
\kern-0.1em
}%
}%
}%
}

\title{Subspace Segmentation by Successive Approximations: A Method for Low-Rank and High-Rank Data with Missing Entries} 

\author{Jo\~ao Carvalho,
        Manuel Marques, Jo\~ao P. Costeira
\thanks{The authors are with the Institute for Systems and Robotics (ISR/IST), Instituto Superior T\'ecnico, Universidade de Lisboa, Portugal.}}

\date{}

\begin{document}

\maketitle

\begin{abstract}
We propose a method to reconstruct and cluster incomplete high-dimensional data lying in a union of low-dimensional subspaces. 
Exploring the sparse representation model, we jointly estimate the missing data while imposing the intrinsic subspace structure.
Since we have a non-convex problem, we propose an iterative method to reconstruct the data and provide a sparse similarity affinity matrix.
This method is robust to initialization and achieves greater reconstruction accuracy than current methods, which dramatically improves clustering performance.
%
Extensive experiments with synthetic and real data show that our approach leads to significant improvements in the reconstruction and segmentation, outperforming current state of the art for both  low and high-rank data.
\end{abstract}

\section{Introduction}
\label{sec:intro}
Linear Subspaces are one of the most  powerful mathematical objects to represent and model high-dimensional data. Machine Learning and especially Computer Vision communities use these tools in a wide variety of algorithms in domains such as classification, structure-from-motion, object recognition and image-segmentation.
However, in all the above areas, observations in real scenarios are incomplete due to self-occlusion, sensor failure or data corruption, to name a few.

In this paper specifically, we address the problem of subspace clustering with missing data by simultaneously completing the data and enforcing a subspace structure.
So, given a set of incomplete high-dimensional points drawn from a union of linear subspaces, we aim to estimate the unknown data and segment the reconstructed data points. 
For example, in Figure \ref{fig:cars_frames}, we focus on the long standing problem of motion segmentation in a scenario with strong occlusion and unreliable feature tracking. The task is to compute the complete feature point trajectories and cluster them according to the motion of the multiple rigid objects in the scene (e.g., cars, background). 
\afterpage{
\begin{figure}[t]
\centering
\includegraphics[width=0.68\textwidth]{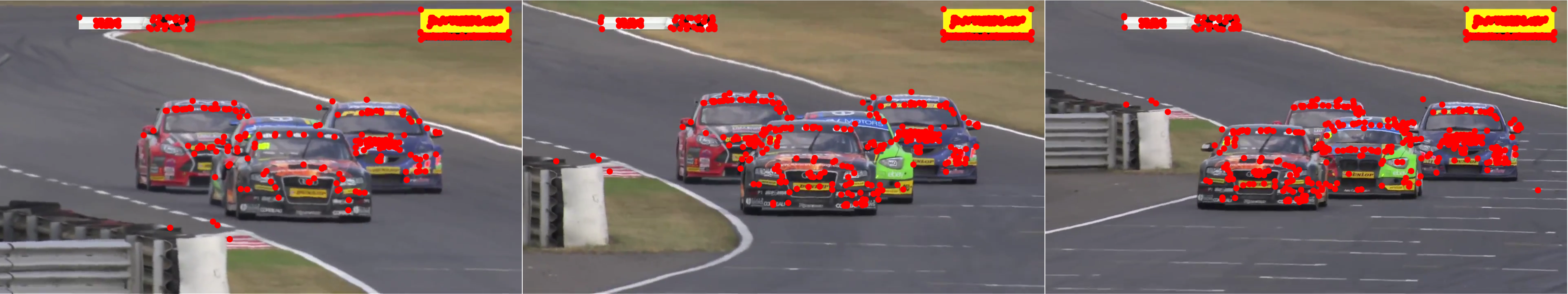}
\caption{Three frames with 4 cars from a racing sequence. In this scenario, cars have similar motion (similar to a single object), and are subject to occlusions. Adding the uncertainty in feature tracking, the motion segmentation in such sequences is a problem difficult to solve.}
\label{fig:cars_frames}
\end{figure}
}

The problem stated above boils down to answering two fundamental questions: 1) how to reconstruct the data from the observed entries and; 2) how to aggregate data such that each cluster lies on a linear subspace. 
Some recent developments attempt to tackle both problems jointly, however, the bulk of published work approaches each of the subproblems separately. For simplicity, we categorize prior work in two main classes: 1) methods that do subspace clustering considering only observed data (\emph{e.g.}, \cite{vidal2015sparse}) and; 2) methods that reconstruct the data assuming it lies on one single subspace and then proceed with the segmentation (\emph{e.g.}, \cite{vidal2004motion}).

We propose a method called \emph{Subspace Segmentation by Successive Approximations} (SSSA), that unifies the reconstruction of the data and the subspace structure of the clusters.
Using a \emph{sparse representation} of the data \cite{candes2006stable, candes2008restricted}, we achieved greater accuracy on the estimates of the missing data, that greatly improved the segmentation.
Extensive experiments prove that this strategy significantly outperforms current state of the art, being able to achieve zero clustering error with up to $70\%$ of data missing in low and high-rank data.


%
\section{Related Work}
\label{sec:related}
%
Most of the state of the art methods for subspace recovery approach the problem without considering the recovery of the missing data, leading to biased and skewed estimates of the subspaces.
Thresholding-based Subspace Clustering (TSC), builds an affinity matrix based on the inner product between incomplete (zero-filled) data points \cite{heckel2015robust}, severely impacting the affinity as the fraction of missing data increases. 
Recently, \cite{vidal2015sparse} proposes the Sparse Subspace Clustering by Entry-Wise-Zero-Fill (SSC-EWZF), generalizing Sparse Subspace Clustering (SSC) \cite{elhamifar2009sparse} by filling the data matrix with zeros in the missing entries and evaluating the error only in the known entries. Also in \cite{vidal2015sparse}, the authors propose the SSC by Column-wise Expectation-based Completion (SSC-CEC), which solves the SSC with an estimated kernel matrix in place of the incomplete data matrix. 
Within this group, several methods assume the rank of the subspaces is known \emph{a priori}.
In \cite{balzano2012k}, the authors initialize $k$ subspaces of known dimensions with the probabilistic farthest insertion and iteratively update the subspaces by minimizing the data projection residuals. 
Based on subset selection, \cite{eriksson2012high} estimates and refines the subspaces of maximum known rank. Missing entries are then estimated by projecting the data onto the assigned subspace. 
Group-sparse subspace clustering (GSSC), proposed in \cite{pimentel2016group}, estimates the subspace bases that best fit the known entries in the data.

An alternative approach is to first estimate the missing data with low-rank matrix completion methods \cite{candes2010matrix}, \cite{cai2010singular}, \cite{wen2012solving} and then segment the data with SSC \cite{elhamifar2009sparse} or low-rank subspace clustering \cite{vidal2014low}. 
Also, in the context of motion segmentation, \cite{vidal2004motion} applies the PowerFactorization method to estimate missing entries after projecting the points onto a low-dimensional space. GPCA is then used to fit linear subspaces to the data \cite{vidal2005generalized}.

Finally, the other approaches jointly recover the missing data and the subspaces. 
In \cite{gruber2004multibody}, the problem is formulated as one of factor analysis, using a mixture of probabilistic PCA to model the data. 
Mixture models are also proposed in \cite{pimentel2014sample} and \cite{pimentel2016group}. The first considers the usual framework of mixture of Gaussians, extending the mixture probabilistic principal component analysers \cite{tipping1999mixtures} to missing data. The second proposes the Mixture Subspace Clustering (MSC), which finds the mixture of data projections (onto each subspace) that best fits the known entries of the data, while minimizing the rank of these projections.
\cite{wen2015sparse} proposes a solution specially tailored for recovering and clustering incomplete images, using total variation regularization for the image restoration. 
More recently, \cite{elhamifar2016high} proposed an extension to SSC where the missing data and coefficients are jointly estimated. Noting that the outer product of a subset of missing entries and coefficients is a rank $1$ matrix, the author proposes the nuclear norm as regularizer. However, it is known that using the nuclear norm to impose a desired rank $r$ may severely distort the $r$ highest singular values, resulting in poor solutions \cite{cabral2013unifying}, \cite{jiang2016robust}.

%

In summary, state of the art methods either use only the observed data, providing biased estimates of the subspaces \cite{heckel2015robust}, \cite{vidal2015sparse}, \cite{balzano2012k}, \cite{eriksson2012high}, \cite{pimentel2016group}, or impose global constraints that assume the data lies on one single subspace, failing to impose the subspace model when reconstructing the data \cite{candes2010matrix}, \cite{cai2010singular}, \cite{wen2012solving}, \cite{vidal2004motion}, or impose inadequate models, failing to jointly recover data and subspaces \cite{gruber2004multibody}, \cite{pimentel2014sample}, \cite{pimentel2016group}, \cite{elhamifar2016high}.
\section{Subspace Segmentation by Successive Approximations} 
\label{sec:method}
In this section, we propose a method to estimate the missing entries of a partially prescribed matrix and to segment the reconstructed data into clusters corresponding to the underlying subspaces.
We build upon the work Sparse Subspace Clustering \cite{elhamifar2009sparse} and explicitly represent the missing data.

Consider $K$ subspaces $\{\mathcal{S}_k\subset \mathbb{R}^D\}_{k=1}^K$, with dimensions $\{d_k<D\}_{k=1}^K$, and let $\{\mathbf{x}_j \in \mathbb{R}^D\}_{j=1}^N$ be the set of $N$ data points lying in the union of the $K$ subspaces (see footnote for notation\footnote{Bold capital letters, $\mathbf{A}$, represent matrices. Bold lower-case letters, $\mathbf{a}$, represent column vectors. Bold lower-case letters with subscript, $\mathbf{a}_i$, represent the $i^{th}$ column of matrix $\mathbf{A}$. Scalars are denoted by non-bold letters, $a$ or $A$. The scalar element in row $i$ and column $j$ of matrix $\mathbf{A}$ is denoted by a non-bold lower-case letter with two subscripts, $a_{ij}$.}).
Denote the data matrix as $\mathbf{X}=[\mathbf{x}_1 \, \mathbf{x}_2  \dots  \mathbf{x}_N] \in \mathbb{R}^{D\times N}$, where each $\mathbf{x}_j$ is a data point. Since only a subset of the entries is observed, we can define $\mathbf{X}$ as
\begin{equation}
    \label{eq:x}
\mathbf{X} = \mathbf{X}_\Omega + \mathbf{X}_{\Omega^{\complement}},
\end{equation}
where $\mathbf{X}_\Omega$ is a matrix with the known entries at $(i,j) \in \Omega$, and zero otherwise. $\mathbf{X}_{\Omega^{\complement}}$ is the matrix with the missing entries in the complementary positions, $(i,j) \in \Omega^\complement$, and zero otherwise. 
To recover the original subspaces, we first estimate the unknown $\mathbf{X}_{\Omega^{\complement}}$ by imposing the subspace model and then cluster the recovered data points.
The missing data is estimated by solving the following optimization problem
\begin{align}
	\min_{\mathbf{X}_{\Omega^{\complement}},\mathbf{C},\mathbf{E},\mathbf{Z}}\quad & \norm{\mathbf{C}}_1 + \lambda_e \norm{\mathbf{E}_\Omega}_1 + \frac{\lambda_z}{2} \norm{\mathbf{Z}_\Omega}_F^2 \label{eq:prob5} \\
	\textup{s.t.}\quad & \mathbf{X}_\Omega + \mathbf{X}_{\Omega^{\complement}} = \left(\mathbf{X}_\Omega + \mathbf{X}_{\Omega^{\complement}}\right)\mathbf{C} + \mathbf{E} + \mathbf{Z} \nonumber \\
	& diag(\mathbf{C}) = 0.\nonumber
\end{align}
The first constraint of \eqref{eq:prob5}, translates the \emph{self-expressiveness} property, which exploits the fact that each data point in a union of subspaces can be represented as a linear or affine combination of other points.
By minimizing the $\ell_1 \text{-norm}$ of the coefficients matrix, $\mathbf{C}$, we favor a \emph{sparse representation} of the data, in which, ideally, a point is represented by a linear combination of few points from its own subspace \cite{candes2006stable,candes2008restricted}. 
The second and third terms in the cost function account for the sparse error, $\mathbf{E}$, and noise, $\mathbf{Z}$, in the known entries, respectively.
For complete data, this model is guaranteed to recover the desired representation when the subspaces are sufficiently separated and data points are well distributed inside the subspaces \cite{elhamifar2013sparse, soltanolkotabi2012geometric, soltanolkotabi2014robust}.

We depart from \cite{vidal2015sparse} by imposing the model to both complete and incomplete data and dealing with outliers, we dramatically improve the recovery of the underlying subspaces, providing a better fit to the known entries of the data.
The problem becomes non-convex because of the product between $\mathbf{X}_{\Omega^{\complement}}$ and $\mathbf{C}$, so, in the next section, we propose an iterative algorithm to solve \eqref{eq:prob5}.
Finally, after solving the optimization problem, we build an affinity matrix from $\mathbf{C}$ and apply spectral clustering to segment the data in $K$ clusters.

\subsection{Solving the Optimization Problem}
First, we note that \eqref{eq:prob5} is convex when one of the variables, $\mathbf{X}_{\Omega^{\complement}}$ or $\mathbf{C}$, is fixed. 
So, similar to \cite{guerreiro2003estimation}, we propose a two step iterative algorithm, in a \emph{Expectation-Maximization} (EM) approach \cite{dempster1977maximum}. On the \emph{E-step} we estimate the missing data given the current model (subspaces). On the \emph{M-step} we estimate the subspaces given the current missing data. This procedure is outlined in Algorithm \ref{alg:method}.
\begin{algorithm}[hbt]
\caption{Subspace Segmentation by Successive Approximations}
\label{alg:method}
\begin{multicols}{2}
\begin{algorithmic}[1]
\Statex \textbf{Input:} $\mathbf{X}$ - Incomplete data matrix
\Statex \qquad\quad $\Omega$ - Set of indexes of known entries
\Statex \qquad\quad $\lambda_e$, $\lambda_z$ - Error trade off parameters
\Statex \qquad\quad $K$ - Number of subspaces
\Statex \textbf{Output:} $\hat{\mathbf{X}}$ - Estimated data matrix
\Statex \qquad\quad\quad\!$\mathbf{C}$ - Coefficients
\Statex \qquad\quad\quad\!\!$labels$ - Point labels
\Statex 
\Statex
\Statex
\State $\mathbf{X}_{\Omega^{\complement}}^{(0)} \leftarrow$ Initialize \label{alg:inix}
\State  $\mathbf{C}^{(0)} \leftarrow$ Fixing $\mathbf{X}_{\Omega^{\complement}}$ as $\mathbf{X}_{\Omega^{\complement}}^{(0)}$, solve \eqref{eq:prob5} \label{alg:inic}
\While{not convergence}
\Statex \hspace{4.8mm} \emph{E-Step}
\State $\mathbf{X}_{\Omega^{\complement}}^{(i+1)} \leftarrow$ $(\mathbf{X}^{(i)}\mathbf{C}^{(i)})_{\Omega^{\complement}}$ \label{alg:estep}
\Statex \hspace{4.8mm} \emph{M-Step} 
\State  $\mathbf{C}^{(i+1)} \leftarrow$ Fixing $\mathbf{X}_{\Omega^{\complement}}$ as $\mathbf{X}_{\Omega^{\complement}}^{(i+1)}$, \label{alg:mstep}
\Statex \hspace{19.5mm} solve \eqref{eq:prob5} 
\EndWhile
\State $labels \gets$ SpectralClustering($K$,$\mathbf{C}$) \label{alg:cl}
\end{algorithmic}
\end{multicols}
\end{algorithm}
The initialization step in line \ref{alg:inix} provides an initial guess for the missing entries, like zeros in all entries, the (feature-wise) mean of the known entries or random initialization.
In line \ref{alg:estep}, we update the unknown entries with its current estimates, $\mathbf{X}_{\Omega^{\complement}}^{(i+1)}=(\mathbf{X}^{(i)}\mathbf{C}^{(i)})_{\Omega^{\complement}}$. 
In the M-step, line \ref{alg:mstep}, we solve \eqref{eq:prob5} but with fixed $\mathbf{X}_{\Omega^\complement}=\mathbf{X}^{(i+1)}_{\Omega^\complement}$.
Finally, in line \ref{alg:cl}, we use the spectral clustering algorithm.

\section{Experiments}
\label{sec:results}
In this section, we assess the performance of our method and compare it to state of the art methods. We consider two metrics: clustering error and reconstruction error, defined as
\begin{align}
    e_c = \frac{\# \text{missclassified points}}{\# \text{points}},  && 
    e_r = \frac{||\hat{\mathbf{X}} - \mathbf{X}||_F}{\norm{\mathbf{X}}_F},
\end{align}
where $\hat{\mathbf{X}}$ and $\mathbf{X}$ are the estimated and true data matrices, respectively, and $\norm{.}_F$ the Frobenius norm.

We compare our algorithm with SSC-EWZF \cite{vidal2015sparse}, MSC \cite{pimentel2016group}, SSC-Lifting \cite{elhamifar2016high}, TSC \cite{heckel2015robust}, the EM algorithm for subspace clustering (EMSC) \cite{pimentel2014sample} and matrix completion \cite{cai2010singular} followed by SSC (MC-SSC).
The parameters of each algorithm were set according to the authors suggestions and in order to minimize the clustering error in a test set.

In our experiments, we initialized the missing entries with the mean or zeros and use spectral clustering \cite{ng2001spectral} with affinity matrix $|\mathbf{C}|+|\mathbf{C}|^T$. We solve \eqref{eq:prob5} with Algorithm \ref{alg:method}. The $\emph{M-step}$ of the algorithm can be solved with the alternating direction method of multipliers (ADMM) \cite{boyd2011distributed}.
\subsection{Synthetic Data}
We draw $N_k$ points per subspace from a union of $K$ random subspaces of dimension $d\ll D$ in $\mathbb{R}^D$. For simplicity, we draw the same number of points $N_k$ per subspace $k \in \{1,\dots,K\}$ and assume all subspaces have dimension $d \leq N_k$.
The missing data is generated by selecting uniformly at random, with probability $\rho$, the set of entries corresponding to missing values, $\Omega^\complement$.

To benchmark the performance of the methods, we evaluate the impact of the rank of the data matrix by running the methods with low-rank and high-rank data. In the low-rank (LR) case we have: $K=3$, $D=50$, $d_k=5$, $N_k=20$. Unless stated otherwise, in the high-rank (HR) case we have $K=10$, $D=80$, $d_k=10$, $N_k=50$. Moreover, we study the impact of the missing rate, ambient space dimension and number of points per subspace in the reconstruction and clustering errors. Finally, we assess the performance of our method when the data is corrupted with noise or outlying entries.

\subsubsection{Reconstruction Error}
\begin{figure}[bht]
\centering
\subfloat[Varying $\rho$ in low-rank.]{\includegraphics[width=0.245\textwidth]{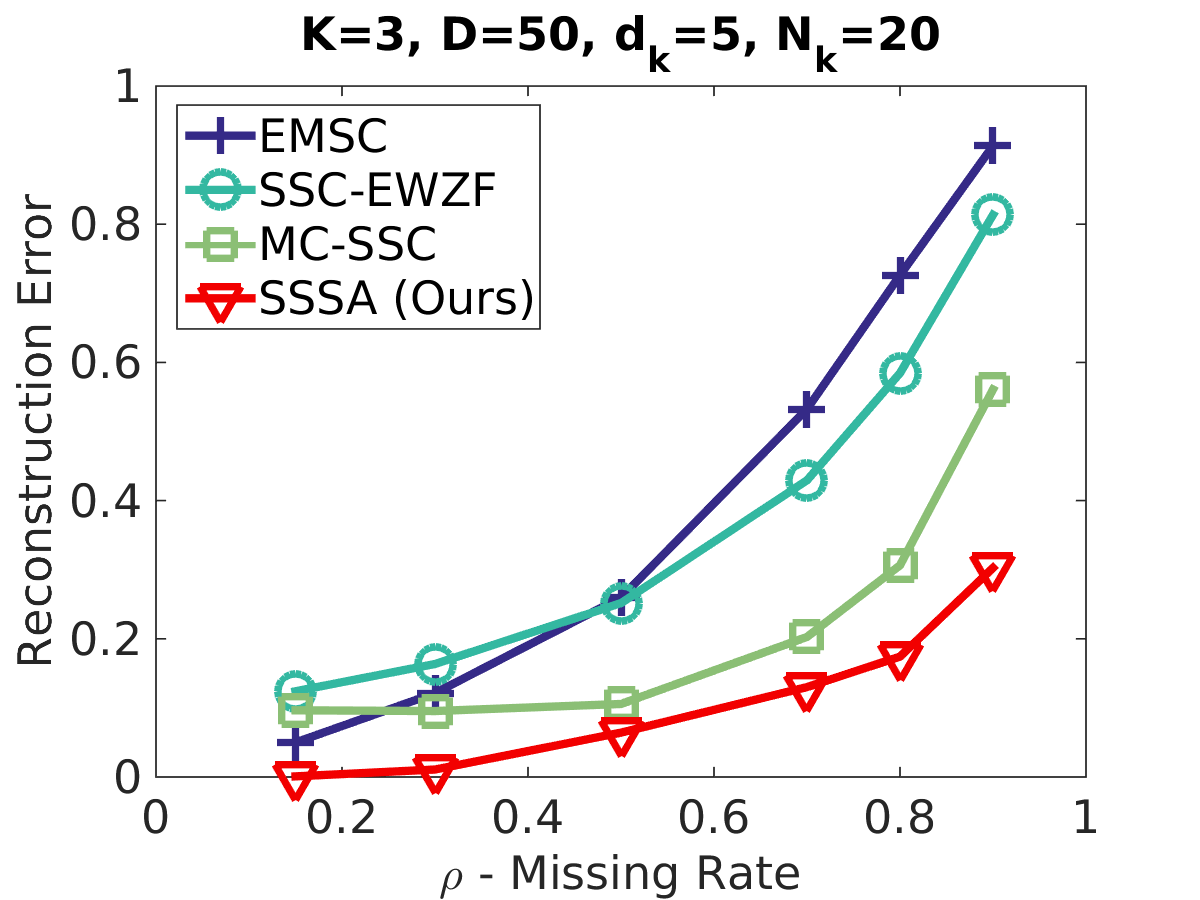}
\label{fig:er_lr_nk20}}
\subfloat[Varying $\rho$ in high-rank.]{\includegraphics[width=0.245\textwidth]{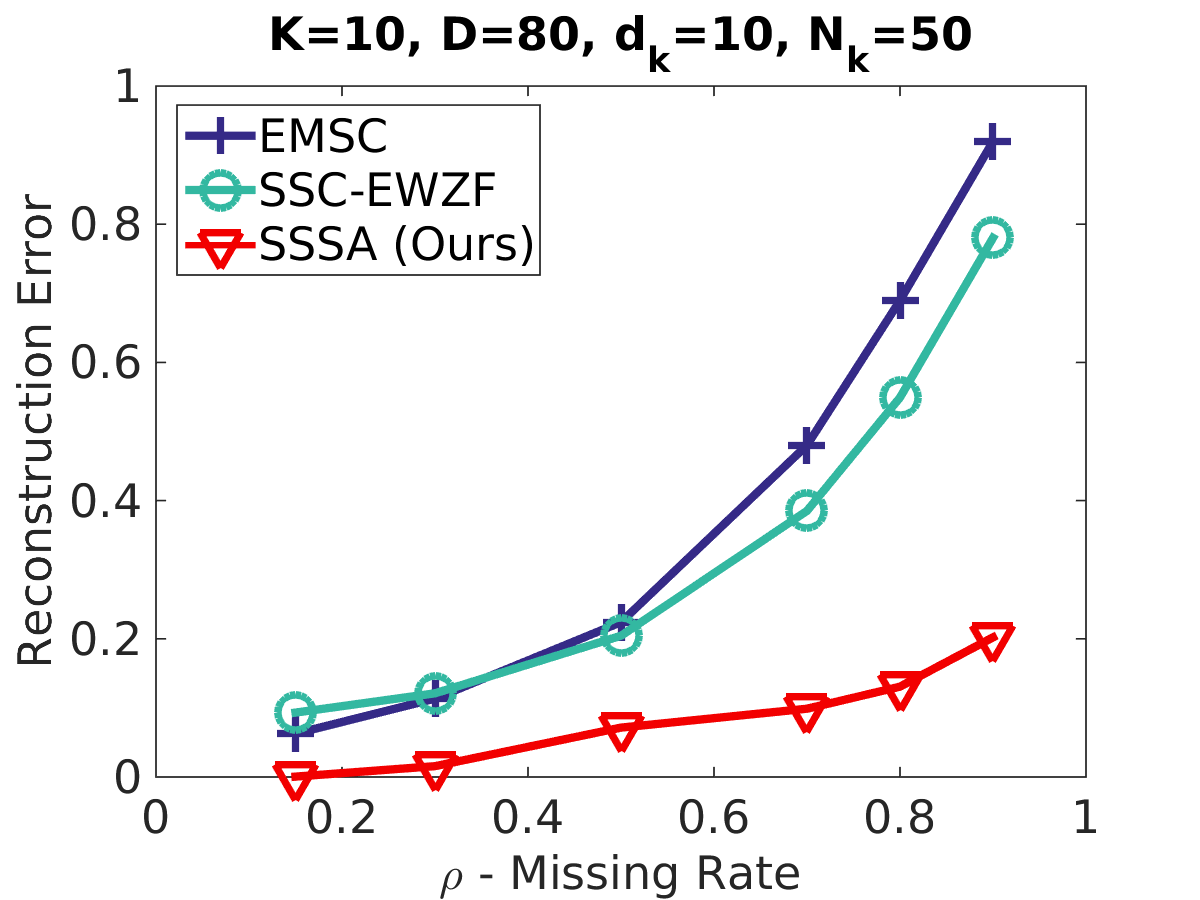}
\label{fig:er_hr_nk50}}
\subfloat[Varying $D$.]{\includegraphics[width=0.245\textwidth]{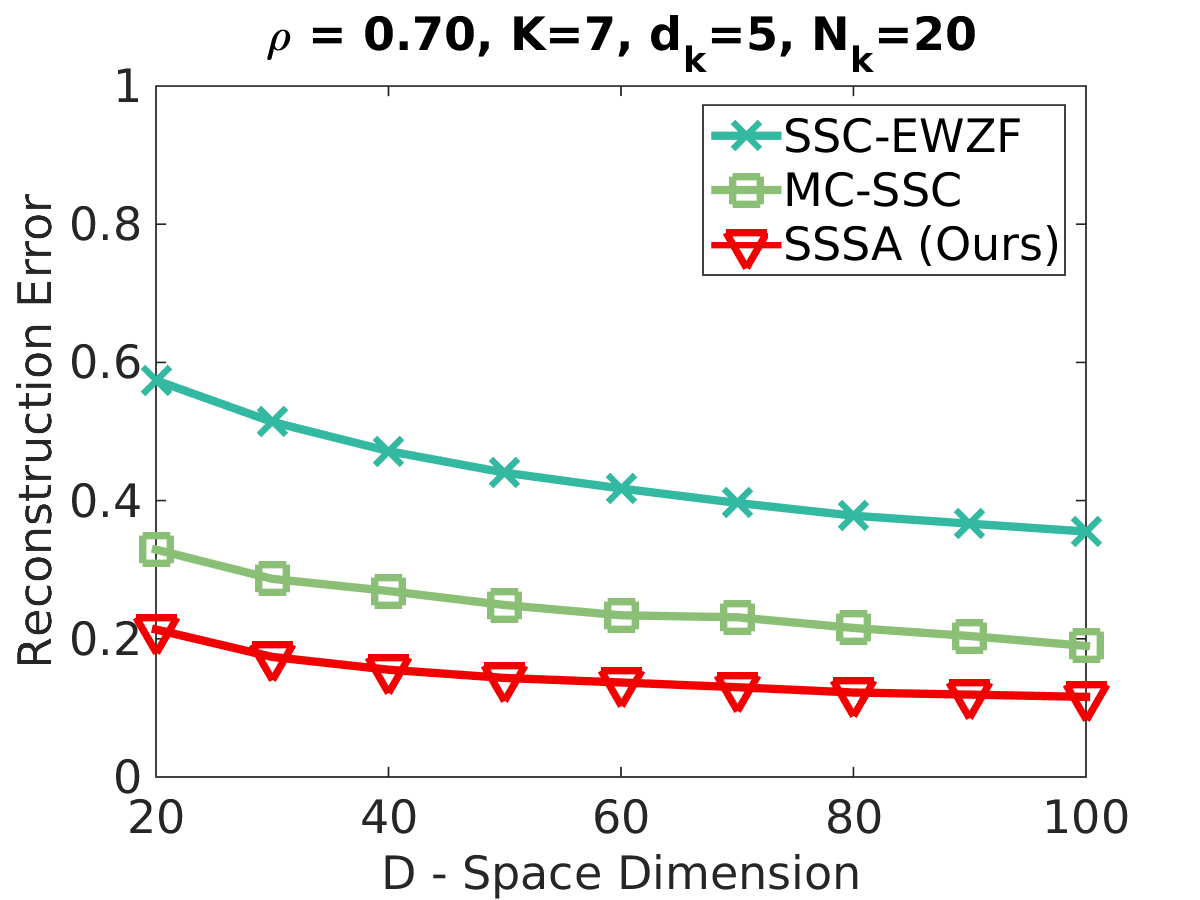}
\label{fig:er_si_070}}
\subfloat[Varying $N_k$.]{\includegraphics[width=0.245\textwidth]{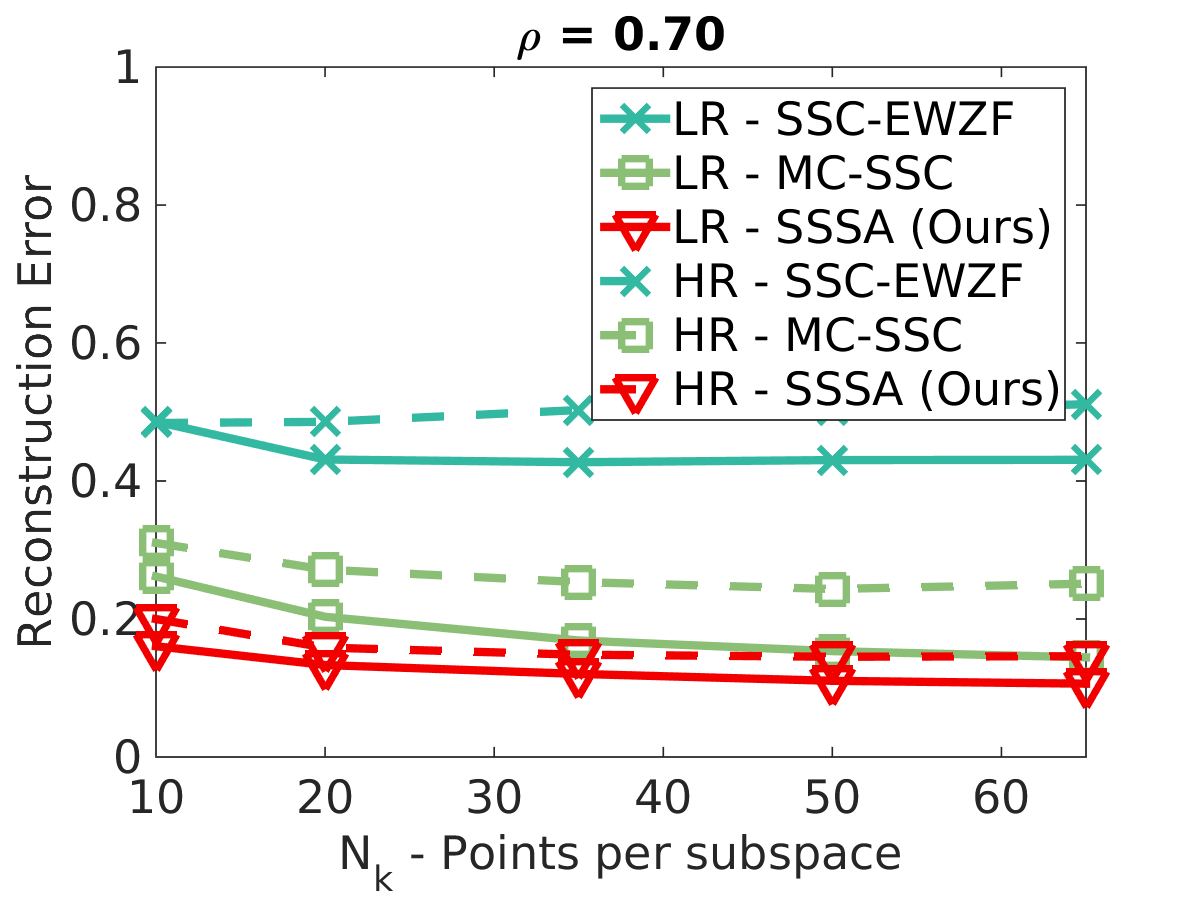}
\label{fig:er_pi_070}}
\caption{Reconstruction error: (a) as a function of missing rate, $\rho$, for LR case; (b) as a function of missing rate, $\rho$, HR case; (c) as a function of the ambient space dimension, $D$, with $\rho=0.70$, $N_k=20$, $K=7$ and $d=5$; (d) as a function of the number of points per subspace, $N_k$, with $\rho=0.70$ for low-rank ($K=3$, $d=5$, $D=50$) and high-rank ($K=7$, $d=5$, $D=35$).} 
\label{fig:rec_err}
\end{figure}
%

Figures \ref{fig:er_lr_nk20}-\ref{fig:er_hr_nk50} show the reconstruction error as a function of the missing probability, $\rho$, for LR and HR cases, respectively. 
In both cases, SSSA outperforms other methods. This is specially noticeable for higher missing rates, where only MC-SSC comes close in low-rank. However, this method is intrinsically inadequate for high-rank data. %

Figure \ref{fig:er_si_070} shows the error versus the ambient space dimension $D$, going from high to low-rank. In this experiment we have $\rho=0.70$, $N_k=20$, $K=7$ and $d=5$. As before, SSSA has significantly lower error, which is crucial in achieving a good subspace segmentation, as we will see in the next section.
With a more favorable case ($\rho=0.30$ and $N_k=100$), SSC-Lifting has higher reconstruction errors, as reported by the author in \cite{elhamifar2016high}. With $D=20$, it reports $e_r \approx 0.495$ and with $D=30$, $e_r \approx 0.36$. Moreover, this method is computationally expensive, as for each missing entry it introduces $N$ new variables, with $N=\sum_k^K N_k$, exponentially increasing the size of the problem as $\rho$ or $N_k$ increase.
%

Figure \ref{fig:er_pi_070} shows the reconstruction error as a function of the number of points per subspace for low-rank ($K=3$, $d=5$, $D=50$) and high-rank ($K=7$, $d=5$, $D=35$) cases, with $\rho=0.70$. Although all methods have approximately constant error as $N_k$ increases, SSSA has lower error in both low and high-rank, with very similar results in both cases.
\subsubsection{Clustering Error}
\begin{figure}[bht]
\centering
\subfloat[Varying $\rho$ in low-rank.]{\includegraphics[width=0.245\textwidth]{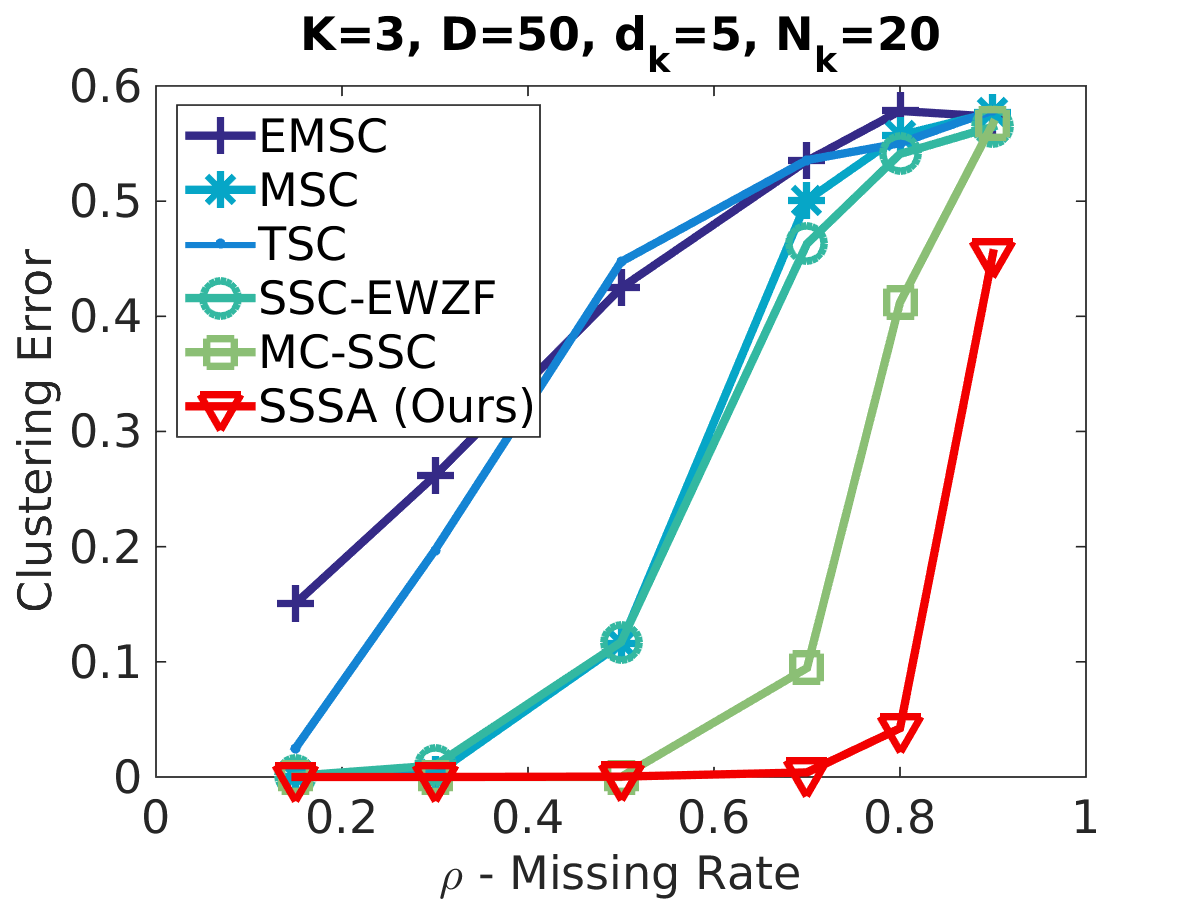}
\label{fig:ec_lr_nk20}}
\subfloat[Varying $\rho$ in high-rank.]{\includegraphics[width=0.245\textwidth]{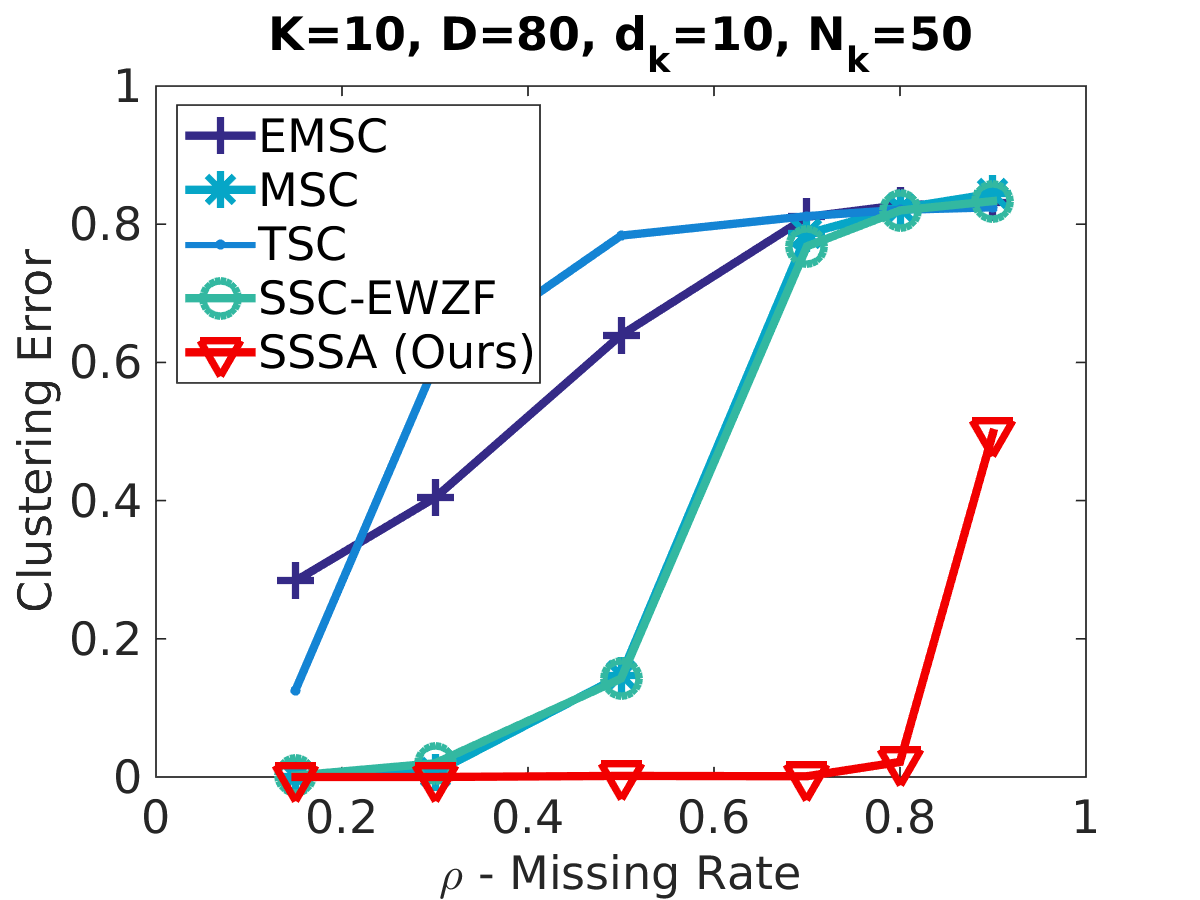}
\label{fig:ec_hr_nk50}}
\subfloat[Varying $D$.]{\includegraphics[width=0.245\textwidth]{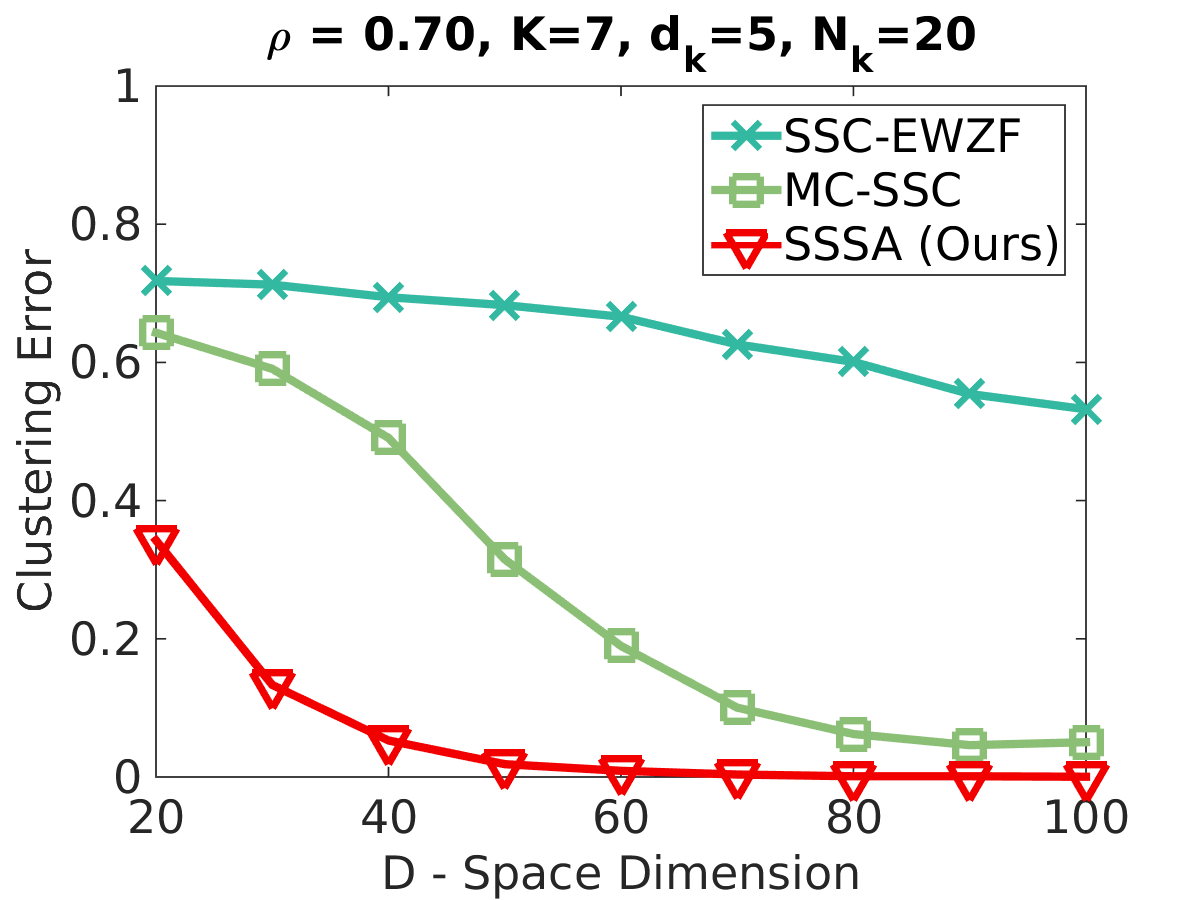}
\label{fig:ec_si_070}}
\subfloat[Varying $N_k$.]{\includegraphics[width=0.245\textwidth]{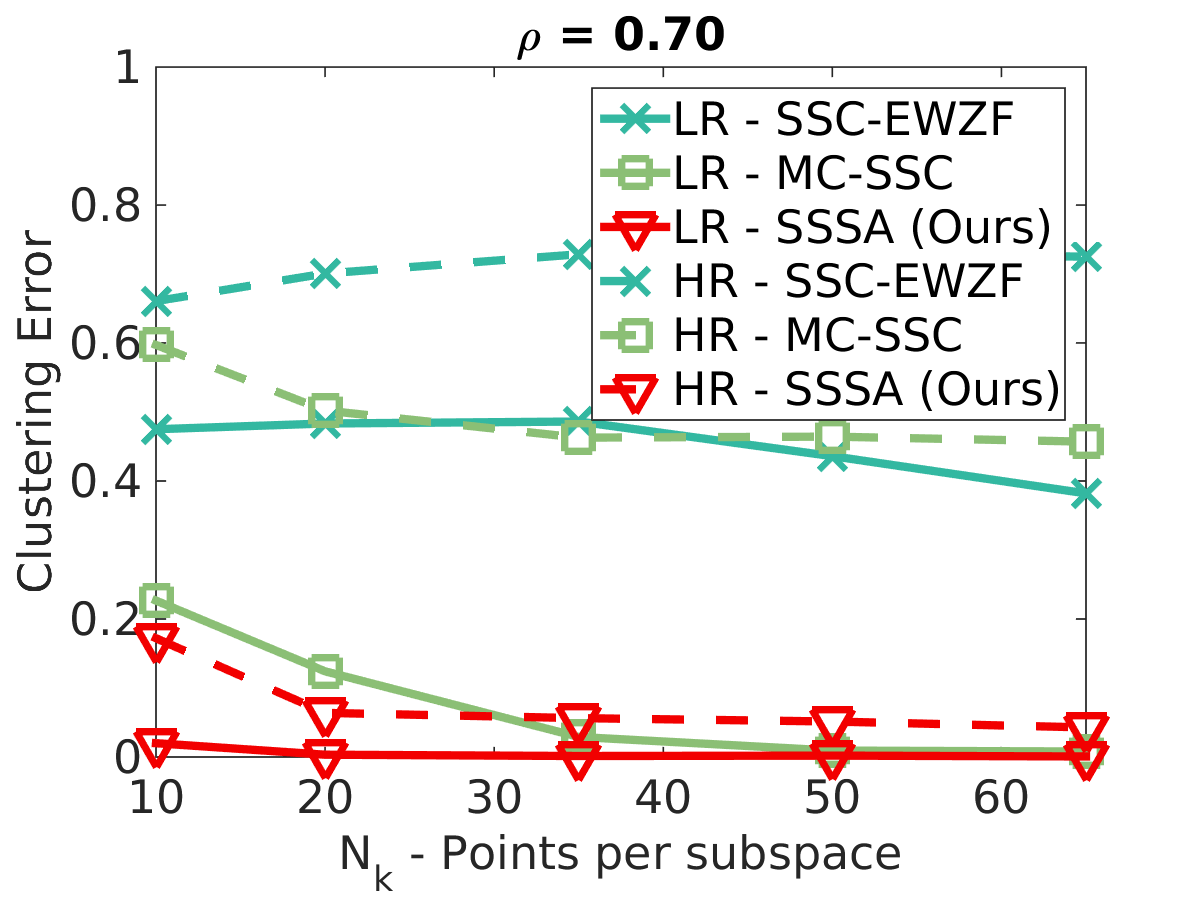}
\label{fig:ec_pi_070}}
\caption{Reconstruction error: (a) as a function of missing rate, $\rho$, for LR case; (b) as a function of missing rate, $\rho$, HR case; (c) as a function of the ambient space dimension, $D$, with $\rho=0.70$, $N_k=20$, $K=7$ and $d=5$; (d) as a function of the number of points per subspace, $N_k$, with $\rho=0.70$ for low-rank ($K=3$, $d=5$, $D=50$) and high-rank ($K=7$, $d=5$, $D=35$).} 
\label{fig:clust_err}
\end{figure}
%
Figures \ref{fig:ec_lr_nk20} and \ref{fig:ec_hr_nk50} show the clustering error as a function of $\rho$ for LR and HR. SSSA has significantly lower errors in the harder cases, with high missing rates.
We should note that SSC-EWZF is equivalent to SSSA first iteration if we initialize the missing entries with zeros and there are no outliers in the data. However, the difference in performance between SSSA and SSC-EWZF highlights the improvement achieved by the iterative approach we propose for the subspace recovery.
%

Figure \ref{fig:ec_si_070} shows the clustering error as a function of $D$ for $\rho=0.70$, $N_k=20$, $K=7$ and $d=5$. The gap between SSSA and other methods reinforces, once again, the dramatic improvement that our method brings over the state of the art methods in the harder cases, \emph{i.e.}, high-rank data and high missing rates.
%

Finally, Figure \ref{fig:ec_pi_070} shows the clustering error versus $N_k$ for the same cases as Figure \ref{fig:er_pi_070}. With $70\%$ missing data, in any scenario, our method has meaningful gains in the clustering performance. MC-SSC error decreases as $N_k$ increases but it still has large errors in the high-rank case.
%
\subsubsection{Noise and Outliers}
\begin{figure}[bht]
\centering
\subfloat[]{\includegraphics[width=0.245\textwidth]{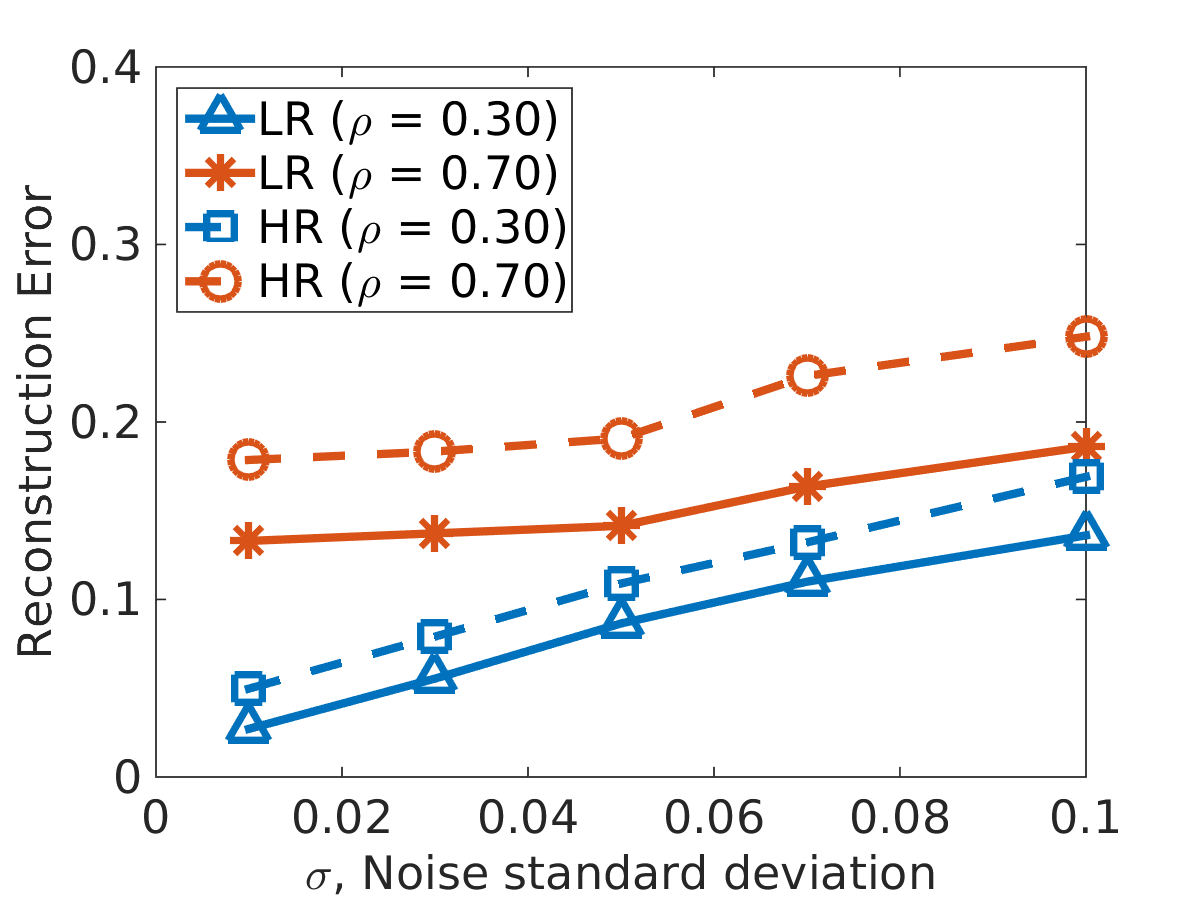}
\label{fig:er_noise}}
\subfloat[]{\includegraphics[width=0.245\textwidth]{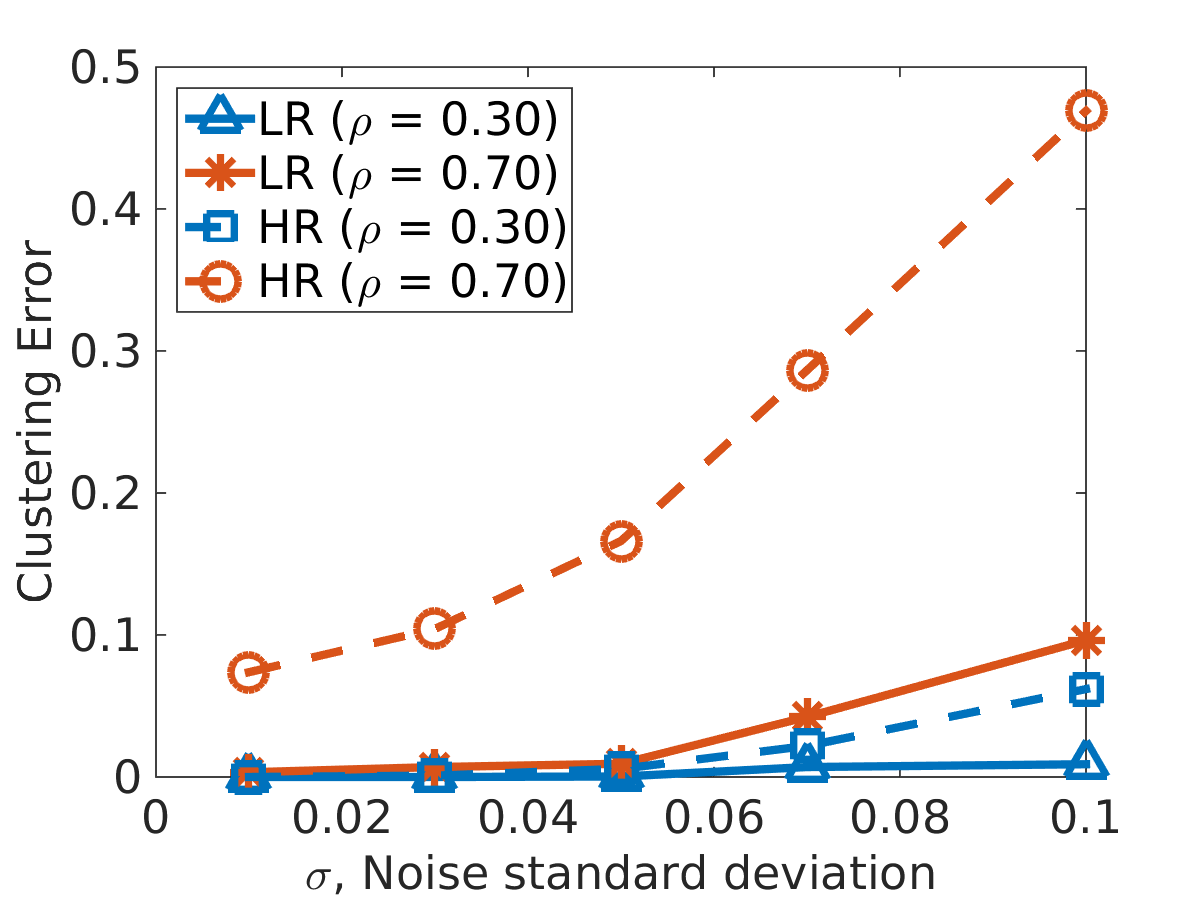}
\label{fig:ec_noise}}
\subfloat[]{\includegraphics[width=0.245\textwidth]{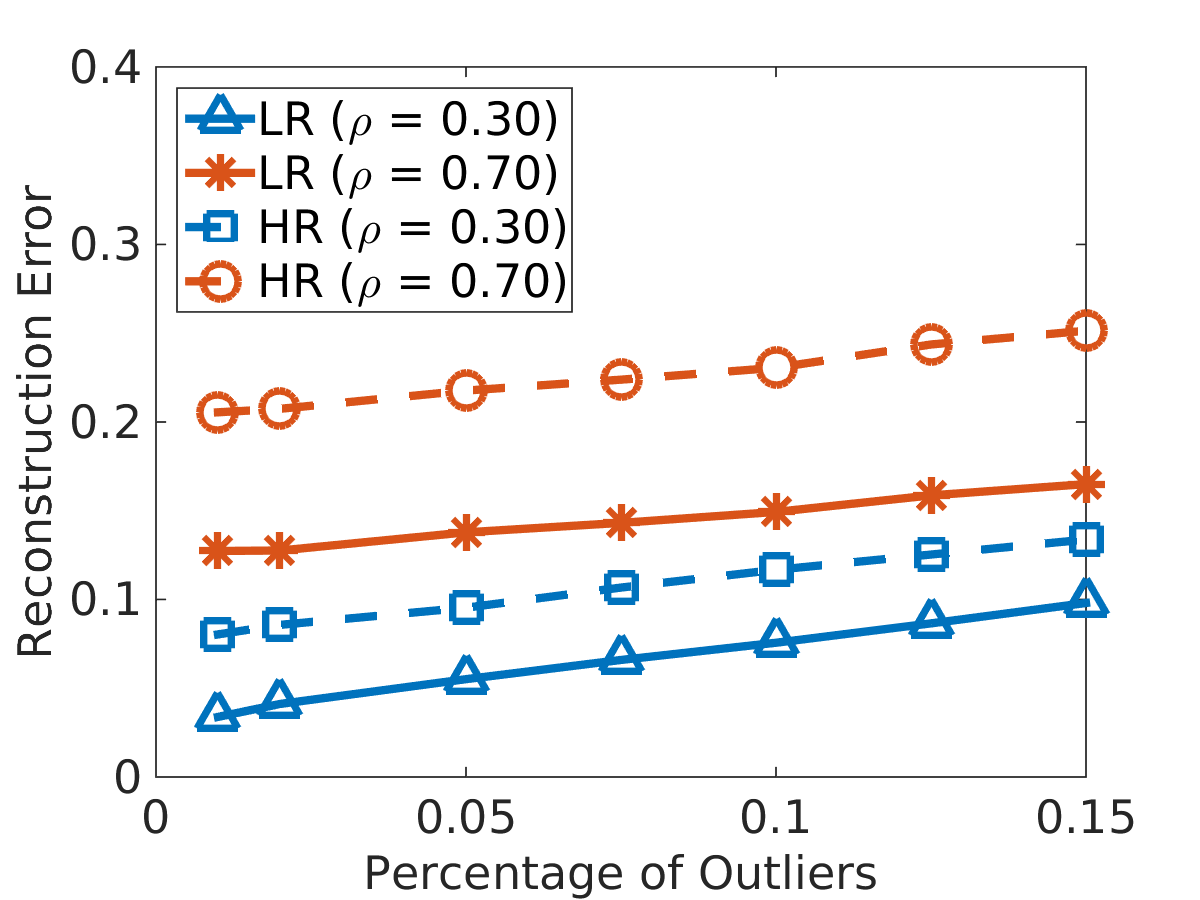}
\label{fig:er_outliers}}
\subfloat[]{\includegraphics[width=0.245\textwidth]{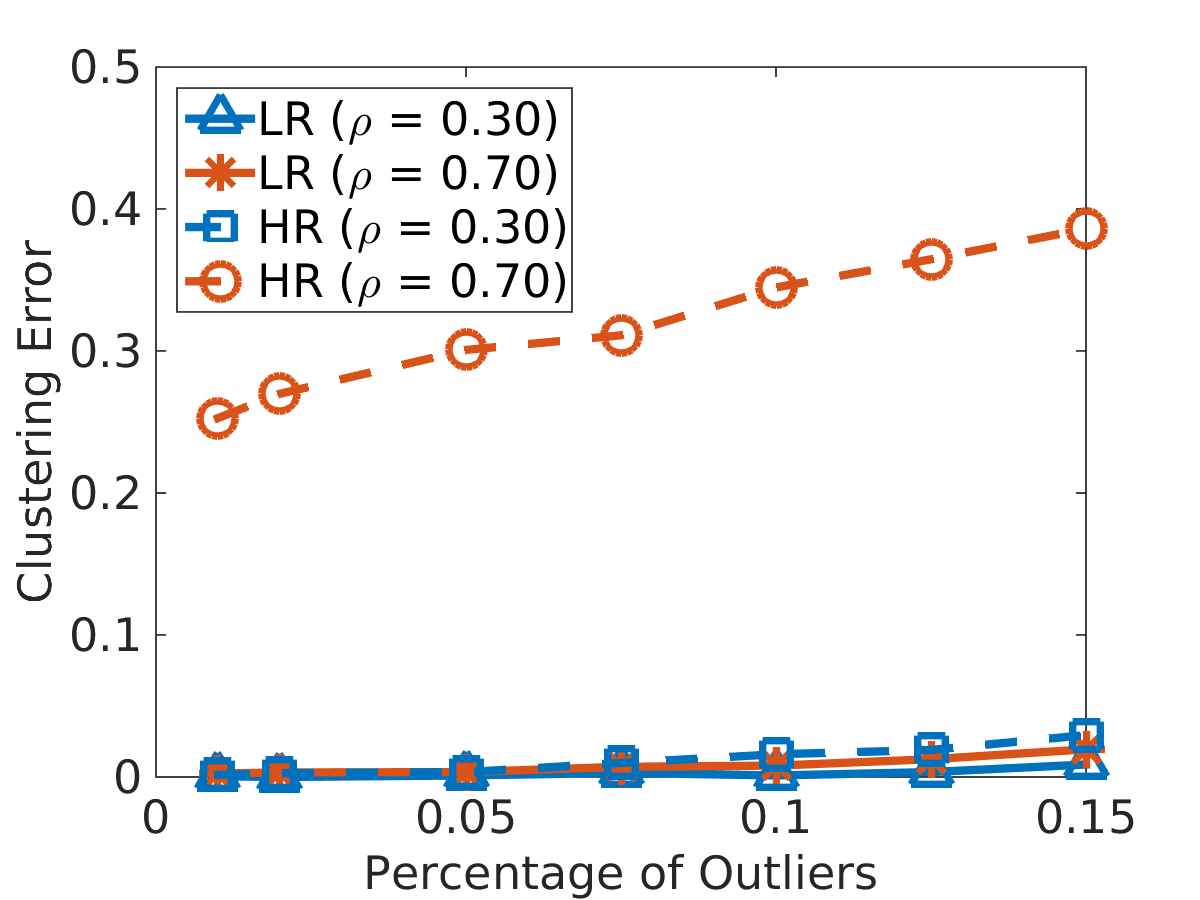}
\label{fig:ec_outliers}}
\caption{Noise and outliers' impact in low-rank (LR) and high-rank (HR) data: (a) and (c) reconstruction error with added noise and outliers, respectively; (b) and (d) clustering error with added noise and outliers, respectively. LR: $K=3$, $d_k=5$,$D=50$, $N_k=50$; HR: $K=7$, $d_k=5$,$D=35$, $N_k=30$; $\rho \in \{0.30, 0.70\}$.}
\label{fig:noise_outliers}
\end{figure}
To assess the robustness of our method to noise and corrupted entries, we generate low-rank and high-rank data (with range $[0,1]$) and add noise or outliers. We add gaussian noise with zero mean and standard deviation $\sigma$. 
To create outlying entries, we generate random values with uniform distribution, amplitude in $[0.2,1]$ and add them to a percentage of entries selected uniformly at random.

Figures \ref{fig:er_noise} and \ref{fig:ec_noise} show, respectively, the reconstruction and clustering errors for added noise, as $\sigma$ increases in $[0.01, 0.10]$. 
In the low-rank case, noise has no significant impact in the clustering performance up to $\sigma=0.10$. For high-rank with $30\%$ missing, the noise effect becomes evident after $\sigma=0.05$, while for $\rho = 0.70$ the effect is much more meaningful.
Figures \ref{fig:er_outliers} and \ref{fig:ec_outliers} show the results for data corrupted with outliers in a percentage of entries going from $1\%$ to $15\%$. 
Although the reconstruction error evolves linearly with the percentage of outliers, its impact is significant. However, despite failing to recover the original data points, the clusters remain the same (even if the subspaces are different). Outliers only have a greater impact in the high-rank case with $\rho=0.70$.
%
\subsection{Applications}
In this section, we evaluate our method when applied to motion segmentation and motion capture problems. We show both qualitative and quantitative results. 
%
\subsubsection{Motion Segmentation}
We consider the problem of motion segmentation, where we aim to cluster trajectories of feature points belonging to multiple objects in $F$ frames of a video sequence. Missing data in this context is frequent due to occlusion and failure in feature detection. 

\begin{figure}[bht]
\centering
\subfloat[Feature points.]{\includegraphics[width=0.98\textwidth]{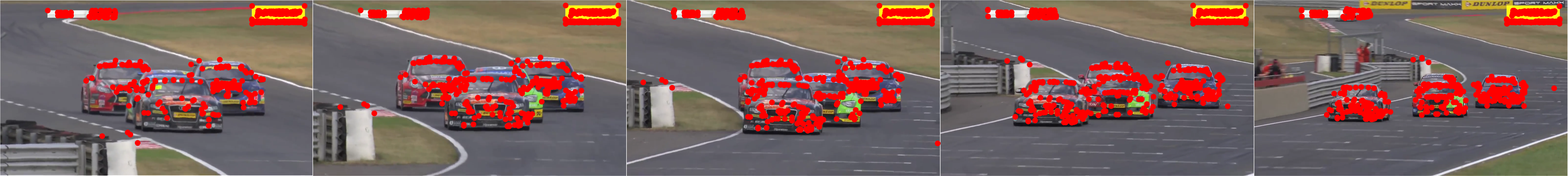}
\label{fig:cars_5fr}}\\
\subfloat[Classified feature points.]{\includegraphics[width=0.98\textwidth]{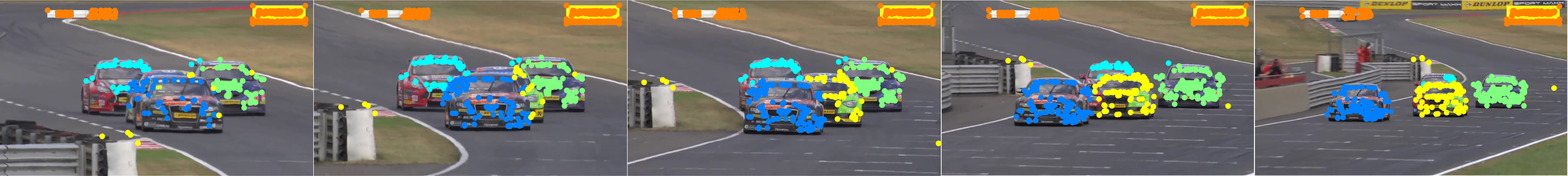}
\label{fig:cars_5fr_cl}}
\caption{Video sequence with 4 racing cars: (a) all feature points in red; (b) classified points, with one color per cluster.}
\label{fig:cars_results}
\end{figure}

First, we consider a sequence of 112 frames with 4 racing cars, from which we show 5 frames in Figure \ref{fig:cars_5fr}. 
In this scenario, all cars aim to ride along the optimal path, having very similar motions (identical to a single object). However, in this sequence, after the curve and going for a long straight segment, they also change motion to try to overtake opponents, leading to occlusions. 
Moreover, feature point trackers are prone to errors and misdetections. For this experiment we use the Kanade-Lucas-Tomasi (KLT) algorithm \cite{lucas1981iterative, shi1994good}.
Despite these challenges, our method succeeds in grouping features from each car, Figure \ref{fig:cars_5fr_cl}.
This experiment shows that our method is able to deal with similar subspaces (motion), occlusions and incomplete feature trajectories.

Next, we assess SSSA with the Hopkins 155 dataset, containing 155 sequences with 2 or 3 objects each. Since this dataset contains only complete trajectories, we add missing data as before, selecting uniformly at random with probability $\rho$ the set of entries corresponding to missing values.
In this dataset the subspaces are affine, therefore, we add the constraint $\mathbf{1^T C=1^T}$ to problem \eqref{eq:prob5}.
\begin{table}[tbh]
\caption{Reconstruction error for all the sequences in Hopkins 155 dataset. For each sequence in the dataset, we generate missing data at random for several $\rho$ with 20 trials each.}
\label{tab:er_hop155}
\centering
\begin{tabular}{lllllllll}
\toprule
$\rho$ & 0.10 & 0.20 & 0.30 & 0.40 & 0.50 & 0.60 & 0.70 & 0.80\\
\midrule
SSC-EWZF &0.070&0.101&0.133&0.183&0.253&0.351&0.481&0.654\\
SSSA     &\textbf{0.005}&\textbf{0.005}&\textbf{0.005}&\textbf{0.006}&\textbf{0.011}&\textbf{0.021}&\textbf{0.039}&\textbf{0.119}\\
\bottomrule
\end{tabular}
\end{table}
\begin{table}[tbh]
\caption{Clustering error for all the sequences in Hopkins 155 dataset as a function of $\rho$. The results for the SSC-Lifting and MC-SSC are as reported in \cite{elhamifar2016high} and \cite{vidal2015sparse}, respectively.}
\label{tab:ec_hop155}
\centering
\begin{tabular}{lllllllll}
\toprule
$\rho$ & 0.10 & 0.20 & 0.30 & 0.40 & 0.50 & 0.60 & 0.70 & 0.80\\
\midrule
SSC-EWZF &0.180&0.204&0.226&0.245&0.257&0.275&0.296&0.318\\
MC-SSC &0.049&0.049&0.049&0.049&0.049&-&-&-\\
SSC-Lifting &0.024&0.025&0.022&0.023&0.028&0.033&\textbf{0.033}&-\\
SSSA     &\textbf{0.016}&\textbf{0.016}&\textbf{0.016}&\textbf{0.018}&\textbf{0.018}&\textbf{0.022}&\textbf{0.033}&\textbf{0.086}\\
\bottomrule
\end{tabular}
\end{table}

Table \ref{tab:er_hop155} shows the average reconstruction error in the 155 sequences for SSSA and SSC-EWZF. Table \ref{tab:ec_hop155} shows the average clustering error for the same sequences. Here, we include SSC-Lifting, as reported by the author in \cite{elhamifar2016high}. 
In our experiments with the Hopkins data set, the MC algorithm did not converge with the parameters suggested by the authors. Thus, the results presented here for MC-SSC are the ones reported in \cite{vidal2015sparse}.
%
%
For all levels of $\rho$, our method achieves the lowest errors.

\begin{figure}[t]
\centering
\includegraphics[width=0.99\textwidth]{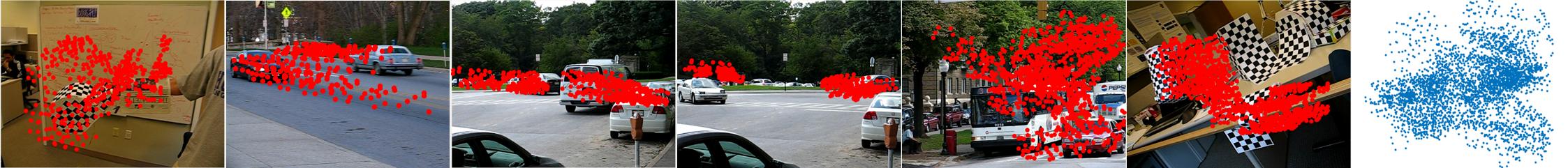}
\caption{First frame of 6 sequences from the Hopkins 155 dataset. We build a data matrix with the trajectories of the objects from these sequences, in a total of 15 objects. The image on the right shows the points from the 15 objects in the same frame.}
\label{fig:hop_hr_gt}
\end{figure}

The data of these sequences are low-rank, since we have 2 or 3 rigid objects and the dimension of each subspaces is at most $d=4$ in $D=2F \geq 28$. 
Therefore, to assess the performance in high-rank, we merge 6 sequences in one data matrix, with a total of 12 objects. Since each sequence has a different number of frames, we subsample the sequences into 10 frames ($D=20$), selecting frames from the original sequence as spread apart as possible.
Figure \ref{fig:hop_hr_gt} shows the first frame of each of these sequences and all the feature points from the $F$ frames. 
The right image plots all points in the same reference frame.
Tables \ref{tab:er_hop_hr} and \ref{tab:ec_hop_hr} show the errors for this high-rank case, where SSSA outperforms, as expected, the SSC-EWZF algorithm.
\begin{table}[tbh]
\caption{Reconstruction error as a function of the missing rate, $\rho$, for a data matrix with 12 objects from 6 sequences from the Hopkins 155 dataset. We report the mean error of 20 trials for each $\rho$.}
\label{tab:er_hop_hr}
\centering
\begin{tabular}{lllllllll}
\toprule
 $\rho$& 0.10 & 0.20 & 0.30 & 0.40 & 0.50 & 0.60 & 0.70 & 0.80\\
\midrule
SSC-EWZF &0.065& 0.145& 0.223& 0.313& 0.415& 0.537& 0.663& 0.791\\
SSSA     &\textbf{0.007}&\textbf{0.009}&\textbf{0.013}&\textbf{0.024}&\textbf{0.064}&\textbf{0.136}&\textbf{0.256}&\textbf{0.427}\\
\bottomrule
\end{tabular}
\end{table}
\begin{table}[H]
\caption{Clustering error as a function of the missing rate, $\rho$, for a data matrix with 12 objects from 6 sequences from the Hopkins 155 dataset. We report the mean error of 20 trials for each level of $\rho$.}
\label{tab:ec_hop_hr}
\centering
\begin{tabular}{lllllllll}
\toprule
 $\rho$& 0.10 & 0.20 & 0.30 & 0.40 & 0.50 & 0.60 & 0.70 & 0.80\\
\midrule
SSC-EWZF &0.236&0.301&0.343&0.430&0.505&0.601&0.680&0.769\\
SSSA     &\textbf{0.008}&\textbf{0.007}&\textbf{0.023}&\textbf{0.090}&\textbf{0.195}&\textbf{0.247}&\textbf{0.334}&\textbf{0.484}\\
\bottomrule
\end{tabular}
\end{table}
\subsubsection{Motion Capture}
We consider the motion capture problem, with the CMU Mocap dataset, where several sensors capture the motion of a human performing various activities. Each point in this dataset corresponds to the measurements of a sensor for a given frame. 
Here, we consider the completion and clustering for a sequence with 7 different activities. Similar to the Hopkins dataset and synthetic experiments, we generate missing data uniformly at random with probability $\rho$. 
%

Table \ref{tab:er_mocap} shows the completion error. The results for SSC-Lifting are as reported by the author for the same sequence \cite{elhamifar2016high}. As before, the improvement achieved by SSSA is specially significant for higher values of $\rho$, where it achieves an error around three times lower than the other methods.

\begin{figure}[h!]
\centering
\subfloat[Without missing data (SSC).]{\includegraphics[width=0.98\textwidth]{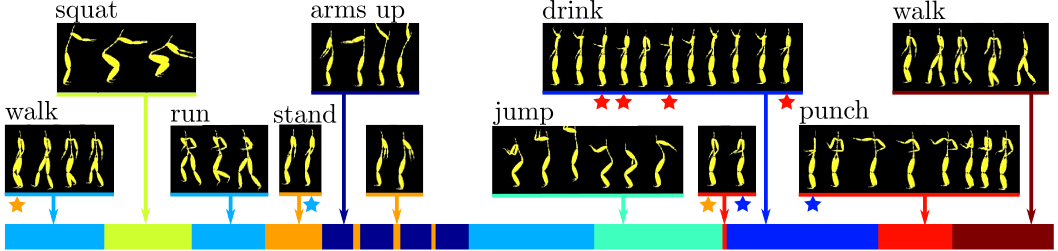}
    \label{fig:mocap_ssc}}\\
\subfloat[SSSA with $50\%$ missing entries.]{\includegraphics[width=0.98\textwidth]{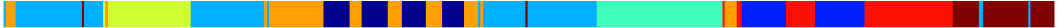}
    \label{fig:mocap_srsa}}
\caption{Clustering labels (one color per cluster) for the CMU Mocap dataset (subject 86, trial 2) and a subset of frames per activity: (a) without missing data (SSC); (b) with $\rho=0.50$ (SSSA).} 
\label{fig:labels_mocap}
\end{figure}

Since the sequence contains smooth transitions between the activities, the labeling is not unique. Therefore, we evaluate the clustering performance qualitatively.
Figure \ref{fig:mocap_ssc} shows the labels for SSC (without missing data) and a set of frames from each activity. 
Without imposing any continuity constraint, as in \cite{tierney2015segmentation}, the clusters are temporally well defined. With $50\%$ missing data, using our method, there are small jumps in some clusters, see Figure \ref{fig:mocap_srsa}.
Colored stars in Figure \ref{fig:mocap_ssc} highlight similar poses in different clusters. 
For example, red stars indicate frames from the \emph{drink} class which are identical to frames in \emph{punch}.
\begin{table}[hbt]
\caption{Reconstruction error for a sequence of 7 activities (walk, squats, run, stretch, jumps, punches, and drinking) from the CMU Mocap dataset. The results for SSC-Lifting are as reported in \cite{elhamifar2016high}} 
\label{tab:er_mocap}
\centering
\begin{tabular}{llllllll}
\toprule
 $\rho$& 0.10 & 0.20 & 0.30 & 0.40 & 0.50 & 0.60 & 0.70\\
\midrule
SSC-EWZF &0.098&0.131&0.180&0.240&0.321&0.420&0.546\\
SSC-Lifting&0.089&0.125&0.170&0.213&0.285&0.430&0.590\\
SSSA     &\textbf{0.032}&\textbf{0.050}&\textbf{0.068}&\textbf{0.087}&\textbf{0.113}&\textbf{0.148}&\textbf{0.189}\\
\bottomrule
\end{tabular}
\end{table}
%

%
\section{Conclusions}
\label{sec:conclusion}
We proposed a method for subspace segmentation with incomplete data lying in a union of affine subspaces. Subspace Segmentation by Successive Approximations is a method that recovers missing entries by exploiting the sparse representation of the data (for both observed and unobserved entries), constraining it to lie in the respective subspace. 
Using an iterative approach, we significantly improve the reconstruction and clustering performance.
Extensive synthetic and real data experiments showed that our method dramatically outperforms current state of the art methods for both low and high-rank problems.



\bibliography{sc-miss-bib}
\bibliographystyle{abbrv}






\end{document}